\documentclass[conference]{IEEEtran}
\IEEEoverridecommandlockouts
\usepackage{amsmath}
\usepackage{amssymb}
\usepackage{amsfonts}
\usepackage{amsthm}
\usepackage{epsfig}
\usepackage{subfigure}
\usepackage{calc}
\usepackage{color}
\usepackage[all]{xy}
\usepackage{bm}
\usepackage{algpseudocode}
\usepackage{algorithm}
\usepackage{enumerate}
\usepackage{cite}
\usepackage{multicol}
\usepackage{graphicx}
\usepackage{stfloats}
\usepackage{bbm}
\usepackage{enumitem}
\usepackage{textcomp}
\usepackage{xcolor}
\usepackage{pseudocode}
\usepackage[lighttt]{lmodern}
\usepackage{listings}
\usepackage[letterpaper, left=1in, right=1in, bottom=1in, top=0.75in]{geometry}
\def\BibTeX{{\rm B\kern-.05em{\sc i\kern-.025em b}\kern-.08em
    T\kern-.1667em\lower.7ex\hbox{E}\kern-.125emX}}

\title{Transfer Learning Based Automatic Model Creation Tool For Resource Constraint Devices}

\makeatletter
\newcommand{\linebreakand}{%
  \end{@IEEEauthorhalign}
  \hfill\mbox{}\par
  \mbox{}\hfill\begin{@IEEEauthorhalign}
}
\makeatother

\author{\IEEEauthorblockN{1\textsuperscript{st} Karthik Bhat}
\IEEEauthorblockA{\textit{Samsung R\&D Institute }\\
Bangalore,India \\
kv.bhat@samsung.com}
\and

\IEEEauthorblockN{2\textsuperscript{nd} Manan Bhandari}
\IEEEauthorblockA{\textit{Samsung R\&D Institute} \\
Bangalore,India \\
m.bhandari@samsung.com}
\and

\IEEEauthorblockN{3\textsuperscript{rd} ChangSeok Oh}
\IEEEauthorblockA{\textit{Samsung Research} \\
Seoul, South Korea \\
seok.oh@samsung.com }
\\\linebreakand

\IEEEauthorblockN{4\textsuperscript{th} Sujin Kim}
\IEEEauthorblockA{\textit{Samsung Research} \\
Seoul, South Korea \\
sjsujin.kim@samsung.com }
\and

\IEEEauthorblockN{5\textsuperscript{th} Jeeho Yoo}
\IEEEauthorblockA{\textit{Samsung Research} \\
Seoul, South Korea \\
jeeho.yoo@samsung.com } 
}

\begin{document}
\IEEEoverridecommandlockouts
\IEEEpubid{\makebox[\columnwidth]{978-1-7281-6916-3/20/\$31.00 \copyright2020 IEEE \hfill} \hspace{\columnsep}\makebox[\columnwidth]{ }}
\maketitle
\IEEEpubidadjcol
\begin{abstract}
With the enhancement of Machine Learning, many tools are being designed to assist developers to easily create their Machine Learning models. In this paper, we propose a novel method for auto creation of such custom models for constraint devices using transfer learning without the need to write any machine learning code. 
We share the architecture of our automatic model creation tool and the CNN Model created by it using pretrained models such as YAMNet and MobileNetV2 as feature extractors.
Finally, we demonstrate accuracy and memory footprint of the model created from the tool by creating an Automatic Image and Audio classifier and report the results of our experiments using Stanford Cars and ESC-50 dataset.
\end{abstract}

\section{Introduction}
Demand for Machine Learning (ML) Systems has been increasing over the past few years and has started to find its application in many domains such as IoT, Medical Care,Energy Management etc.\cite{reddy2018review, dou2019applications,rashid2019machine}. But even with the increase in the success of this concept, developers and business struggle to create and deploy their ML models due to the high learning curve involved in designing these models. This limits its use to a smaller community of data scientists and engineers. Designing a ML Model requires specialized skill set and can be challenging for new developers and businesses. Hence, there has been a growing demand to create tools and systems that can automatically generate ML models for custom user data without much effort or knowledge of Machine Learning.

Concepts such as AutoML \cite{cai2019automl, tuggener2019automated} are already becoming popular to address the above challenges. A survey of state of the art for AutoML can be found at \cite{he2019automl}. Many tools such as CreateML by Apple and Cloud AutoML by Google have been released to address some of the above concerns. Though these tools are very helpful, they have their own limitations. In this paper, we explore a novel method to create an automatic model creation tool using transfer learning that is supported across multiple operating systems and further optimize the model created by our tool to execute on constraint devices seamlessly overcoming the limitations of other tools. 

\subsection{Transfer Learning}
Transfer Learning is a concept in machine learning where the knowledge that is learnt while training any model is applied to another similar model. Transfer learning frameworks and terms like domain and task are summarized in \cite{pan2009survey}. In practice, training any convolutional network from scratch requires a lot of time and resources. Therefore, generally a pretrained convolutional network on a very large dataset with different classes is used either in the initialization or as a feature extractor for a particular task. Major transfer learning scenarios are ConvNet as a fixed feature extractor, fine tuning of the ConvNet and pretrained models.

Technical growth, learning and applications in daily life can be done via images and audios which are prominent in every field like IoT, chatbots, self-driving cars etc. One of the main categories in neural network to do image recognition, image classification, audio classification etc. is convolutional neural network (CNN or ConvNet). A CNN comprises of an input layer, multiple hidden layers and an output layer. The hidden layers mainly consist of many convolutional layers with filters, pooling, fully connected layers. The final convolution mostly involves back propagation in order to weight the end product more accurately \cite{albawi2017understanding}.

\subsection{Image and Audio Classification}
Image classification refers to a computer vision process where the learning model takes an image as the input and outputs to which class it belongs to. The CNN (having multiple convolutional layers, ReLU layer, pooling layers, fully connected layer) takes an input image in the form of pixels and extract features for that image. This has many applications like classification, similarity between images etc. Generally, pretrained CNN models like VGG16, VGG19, MobileNet are used for large image dataset in transfer learning. For classification, support vector machine (SVM), OVR-SVMs, KNN (K-Nearest Neighbors) algorithm can be used. Some of the methods to do image classification are described in \cite{shaha2018transfer}, \cite{beaula2016comparative}, \cite{lee2018image} etc.

Similarly, audio classification refers to a process where audio or sounds from different sources are classified using CNN. It has many important applications like issuing commands to IoT devices, audio control for online gaming etc. Some of the methods to do audio classification are described in \cite{guo2003content}, \cite{kim2019comparison} etc.

Broadly, the audio classification takes place by converting the audio input into an image and applying image classification on it. This can be done by constructing a spectrogram of the audio. Spectrogram is simply a plot of frequency versus time of the audio signal. Spectrograms represent the frequency information of time chunks stringed together as colored vertical bars in the audio as colors are in an image. These are basically two-dimensional graphs, with a third dimension represented by colors.

\section{Related Work}
Image and audio classification has found enormous applications in day-to-day life. Audio classification can be used in surveillance systems that automatically detects incidents like car crashes, tyre skidding etc. It can also be used in user authentication, user family authentication, language detection etc. Some of the already implemented work includes urban sound classification \cite{lezhenin2019urban}, getting information from news video \cite{song2009feature} and many more.

Apart from this, there can be many applications of such classifications for which automated tools are being developed to provide user with a smooth procedure to train and classify their input. Create ML is one such framework introduced by Apple to build the ML models for classification using just drag and drop method. Another such framework is a cloud-based platform provided by Google called Cloud AutoML which is a simple graphical user interface tool to train, evaluate, improve, and deploy models based on the data. 

\section{Our Contribution}
In this paper, we  explore a novel method to create a cross platform Automatic ML model creation tool using Transfer Learning. We share the CNN architecture of models created by our tool for image and audio classification and share their on device accuracy and memory footprint.

For Image classification, we evaluate VGG16, VGG19 and MobileNetV2 models as feature extractors and compare the performance of these models with new user provided dataset. We achieve $94.1\%$ accuracy in Image Classification on the subset of Stanford Car Dataset \cite{KrauseStarkDengFei-Fei_3DRR2013}. For Audio classification, we evaluate the YAMNet Model as feature extractor. YAMNet is a pretrained deep net that predicts 521 audio classes based on the AudioSet-YouTube corpus \cite{gemmeke2017audio}, and employs the MobilenetV1 depth-wise separable convolution architecture. We achieve $98.4\%$ in Audio Classification on subset of ESC-50 \cite{DVN/YDEPUT_2015} Environment Sounds Dataset. In both cases we were able to reduce the model size by $\sim74\%$.

\section{Automatic Model Creation Tool Workflow}
In this paper, we explain the architecture of our tool and how we use transfer learning to automatically extract features and train a custom model for user dataset on CPU.
The tool currently supports model creation for image and audio classification.

The tool consists of frontend which is created using Typescript and Electron Framework and the backend CLI (Command Line Interface) which creates the model automatically based on the data provided by the user.

In this section, we explore the backend part of the tool and explain the architecture of the same.
We can divide the tasks performed by the backend tool into 4 parts :
\begin{itemize}
    \item Automatic Data Preparation
    \item Automatic Feature Extraction
    \item Model Training and Evaluation
    \item Model Compression for Constraint Devices
\end{itemize}

We further explain how each of the above steps works in the tool by creating an Image and Audio Classifier as explained in the next section.

\subsection{AUTOMATIC IMAGE CLASSIFIER}
For Automatic Image Classifier, we use MobileNetV2 model for feature extraction.
The steps followed by the tool to create an automatic Image classifier is as below-

\subsubsection{Data Set}
\label{sec411}
We use a subset of Stanford Cars data set. This data set consists of 16,185 images of 196 classes of cars. We take a subset of this data set consisting of 'AM General Hummer SUV 2000', 'BMW 3 Series Wagon 2012', 'Chevrolet Impala Sedan 2007', 'Hyundai Elantra Touring Hatchback 2012' and 'Rolls-Royce Ghost Sedan 2012'. The above data is split into training and validation set.

\subsubsection{Automatic Data Preparation}
The above training and validation data is fed to the tool. The tool preprocesses this data and prepares it as expected by MobileNetV2 model. We perform the following operations on the input Image Data:
\begin{itemize}
    \item Input Images are resized to 224 X 224
    \item Re-scale the RGB coefficients in the range 0-1 by scaling with a 1/255.
    \item Apply Data Augmentation such as Horizontal Flip, Rotation and Zooming. This feature can be configured in tool by user as per the dataset.
\end{itemize}

The processed input is sent to the next phase for feature extraction.

\subsubsection{Feature Extraction}
We load the MobileNetV2 model with pretrained weights and remove the last layer corresponding to prediction. 
We then pass the processed data to extract feature tensors using the MobileNetV2 model.
The extracted features are of shape (samplecount, 7, 7, 1280).

\begin{pseudocode}{FeatureExtractor}{weights}
\PROCEDURE{extractFeatures}{dir,sampleCount}
convBase \GETS MobileNetV2(weights)\\
batchSize \GETS 32 \\ \\

datagen \GETS ImageDataGenerator(rescale)\\
features \GETS np.zeros(shape)\\
labels \GETS np.zeros(shape)\\ \\

generator \GETS datagen.flowFromDir(dir)\\ \\

i \GETS 0\\ 
\FOR {inputBatch, labelsBatch} \in generator \\
\BEGIN
featuresBatch \GETS convBase.predict(inputBatch) \\
\COMMENT{update the feature array} \\
\COMMENT{update the labels array} \\
i \GETS i+1 \\
\IF i*batchSize >= sampleCount \THEN
break\\
\END \\ \\
\RETURN{features,labels}
\ENDPROCEDURE
\end{pseudocode}

\begin{figure}[!h]
    \begin{center}
    \includegraphics[scale=0.5]{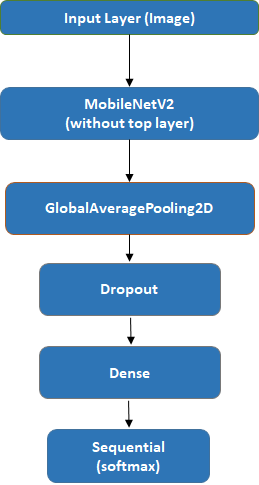}
    \caption{Image Classifier Model Architecture}
    \label{image_classifier_arch}
    \end{center}
\end{figure}

\subsubsection{Model Designing and Training}
The output of the feature extraction phase is then fed to the new model. The model consists of an input layer which accepts the extracted features and adds a GlobalAveragePooling2D, a dropout to prevent overfitting and a dense layer with softmax activation function for classification. Model architecture for the same is described in Figure \ref{image_classifier_arch}.

\begin{figure}[h]
    \centering
    \subfigure{{\includegraphics[scale=0.28]{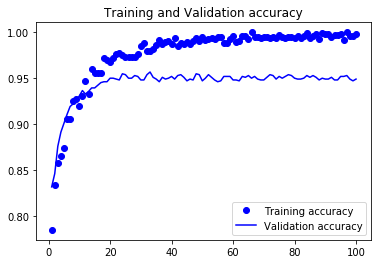} }}%
    \qquad
    \subfigure{{\includegraphics[scale=0.28]{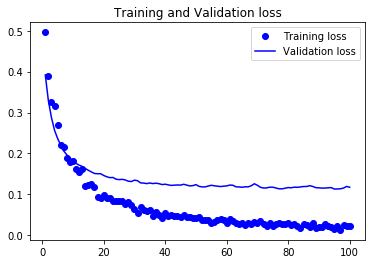}}}%
    \caption{Training/Validation accuracy and loss on car classifier with MobileNetV2}%
    \label{image_classifier_accuracy}
\end{figure}

\begin{figure}[!h]
    \begin{center}
    \includegraphics[scale=0.43]{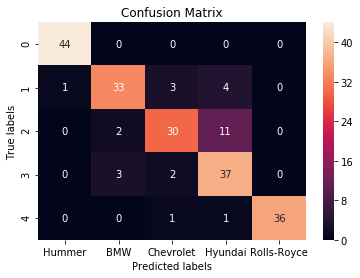}
    \caption{Confusion Matrix for Car Classification}
    \label{image_classifier_confusion}
    \end{center}
\end{figure}

We fit the extracted features of the test and validation data into the above model and create a custom model for our dataset. Figure \ref{image_classifier_accuracy} and \ref{image_classifier_confusion} represents accuracy/loss and confusion matrix obtained with MobileNetV2 as feature extractor respectively.

\subsubsection{Model Compression and Inference}
We club the MobileNetV2 model (without the classifier) and our trained classifier model to create the full model for on device inference. 
Before we flash the model onto device we optimize the model for size by applying post-training quantization such as weight quantization in which we quantify the weights from floating point to 8-bits of precision. At inference, weights are converted back from 8-bits of precision to floating point and computed using floating-point kernels.

We are able to compress the model size by around 74\% with an accuracy of around $94.1\%$  for the image classifier created by the above tool.

\subsection{AUTOMATIC AUDIO CLASSIFIER} 
For Automatic Audio Classifier Model creation, we use YAMNet model for feature extraction.
YAMNet is a pretrained deep net based on the AudioSet-YouTube corpus and employees MobilenetV1 depth wise separable convolution architecture.

The extracted features from YAMNet is then fed as input for our model during training.
We create a environment sound classifier using our tool that can accurately classify musical instruments such as guitar, piano and flute.

\subsubsection{Data Set}
\label{sec421}
We use a subset of ESC-50 Environment sound dataset for training our classifier.
The data set consists of 75 audio files of 5 classes (bell, clap, engine, rain and wind) which is used for training. The validation data consists of 25 audio files of the above classes.

\subsubsection{Automatic Data Preparation}
The tool preprocesses the training and validation data and prepares it as expected by YAMNet model.
We perform the following operations of the input audio data :
\begin{itemize}
    \item Audio is re-sampled and converted to 16 kHz mono channel.
    \item Frames below a threshold are removed to prevent model from training on silent audio frames.
    \item A spectrogram is created using magnitudes of the STFT of window size of 25 ms and window hop of 10 ms.
    \item A mel spectrogram is computed by mapping to 64 mel bins covering the range 125-7500 Hz.
    \item Finally a log mel spectrogram is computed and 50\% overlapping frames are passed to YAMNet model for feature extraction.
\end{itemize}

\subsubsection{Feature Extraction}
In this phase, we preload the trained weights of the YAMNet model and remove the last 3 layers corresponding to the prediction layer of the YAMNet model.
We then pass each of the processed training and validation data from previous step to the YAMNet model and extract the feature tensors.
The extracted features are tensor of shape (samplecount, 3, 2, 1024).

\subsubsection{Model Designing and Training}
\begin{figure}[!h]
    \centering
    \includegraphics[scale=0.5]{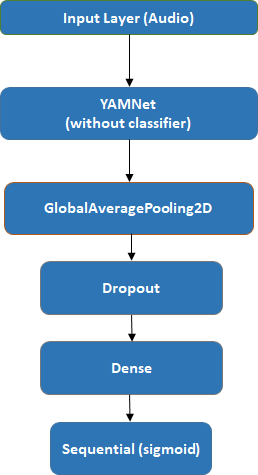}
    \caption{Audio Classifier Model}
    \label{flowchartone}
\end{figure}

The output of the feature extraction phase is then fed to the new model. 
The model consists of an input layer which accepts the extracted features and adds a GlobalAveragePooling2D and a dense layer with sigmoid activation function. Model architecture for the same is described in Figure \ref{flowchartone}. 

\begin{figure}[h]
    \centering
    \subfigure{{\includegraphics[scale=0.27]{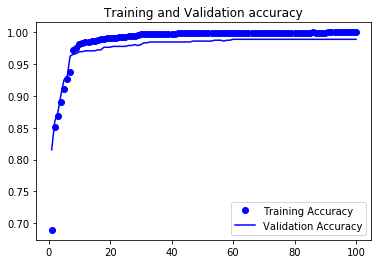} }}%
    \qquad
    \subfigure{{\includegraphics[scale=0.27]{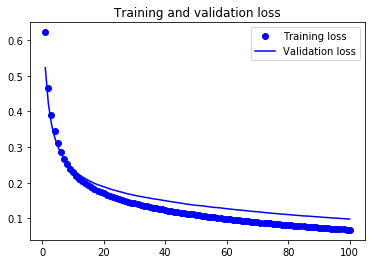}}}%
    \caption{Training/Validation accuracy and loss on environment sounds data}%
    \label{yamnet_env_accuracy}
\end{figure}

\begin{figure}[!h]
    \centering
    \includegraphics[scale=0.4]{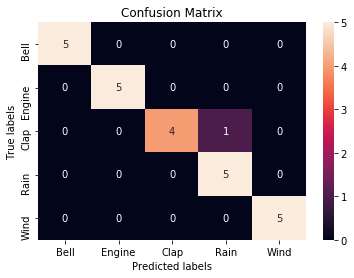}
    \caption{Confusion Matrix for Environment Sounds Classifier}
    \label{yamnet_confusion}
\end{figure}

We fit the extracted features of the test and validation data into the above model and create a custom model for our dataset. Figure \ref{yamnet_env_accuracy} and Figure \ref{yamnet_confusion} represents accuracy/loss and confusion matrix respectively, obtained on the training and validation set with YAMNet model.  

We can see that the model converges quickly and the training and validation loss continue to decrease.

\subsubsection{Model Compression and Inference}
Finally, we club the YAMNet model (without the classifier) and our trained classifier model to create the full model for on device inference. We optimize the model for size as described in the Image Classifier Model Compression and flash it into device for inference. We are able to compress the model size by around 74\%.

We were able to achieve $98.4\%$ accuracy on device for the audio classifier created by the above tool.

\section{Experiments and Results}
Before concluding on the default feature extractor as MobileNetV2 for images and YAMNet for audio, we compare different models and their accuracy and performance as feature extractor.

\begin{figure*}[t]

    \centering
    \subfigure{{\includegraphics[scale=0.20]{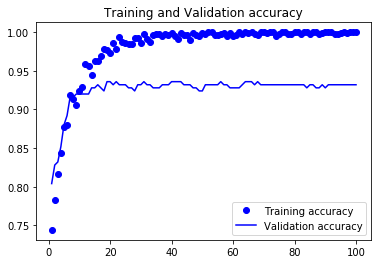}}}%
    \qquad
    \subfigure{{\includegraphics[scale=0.20]{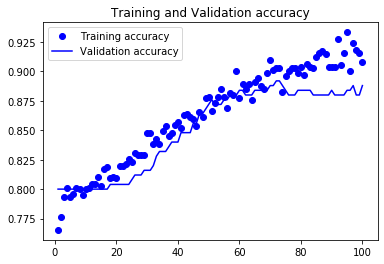}}}%
    \qquad
    \subfigure{{\includegraphics[scale=0.20]{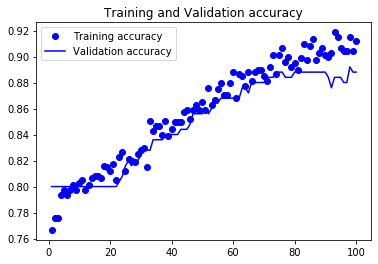}}}%
    \caption{MobileNetV2, VGG16 and VGG19 as feature extractors on Flowers Dataset respectively}
    \label{fig13}

    \centering
    \subfigure{{\includegraphics[scale=0.20]{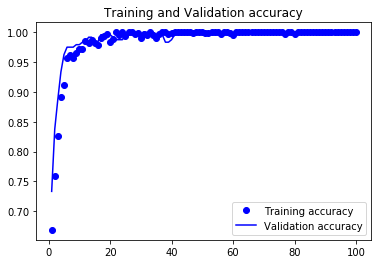}}}%
    \qquad
    \subfigure{{\includegraphics[scale=0.20]{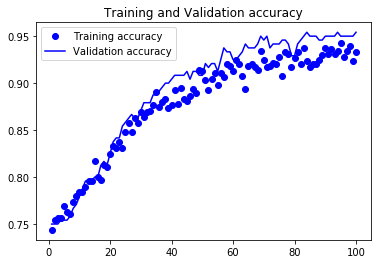}}}%
    \qquad
    \subfigure{{\includegraphics[scale=0.20]{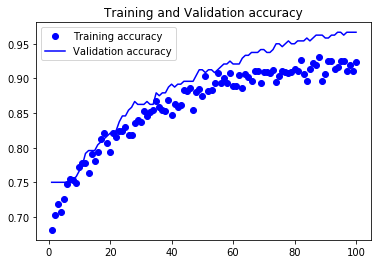}}}%
    \caption{MobileNetV2, VGG16 and VGG19 as feature extractors on Food11 Dataset respectively}
    \label{fig14}
    
    \centering
    \subfigure{{\includegraphics[scale=0.20]{images/accuracy_imagenet}}}%
    \qquad
    \subfigure{{\includegraphics[scale=0.20]{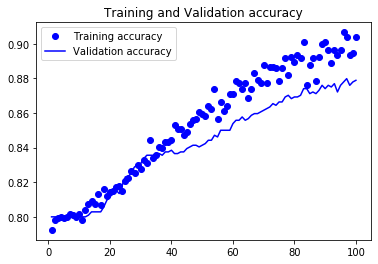}}}%
    \qquad
    \subfigure{{\includegraphics[scale=0.20]{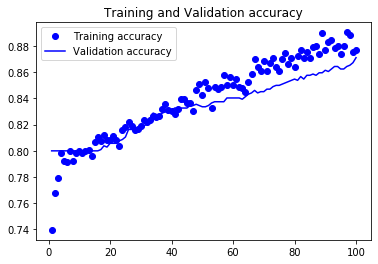}}}%
    \caption{MobileNetV2, VGG16 and VGG19 as feature extractors on Standford Cars Dataset respectively}
    \label{fig15}
  
\end{figure*}

\subsection{Automatic Image Classification}
In case of Image Classification, we compared multiple models such as VGG16, VGG19 and MobileNetV2 on various datasets.
Below are the results with subset of Images from Standford Car dataset, Food-11 dataset and Kaggle Flowers dataset.

{Standford Car Dataset}:
The data is same as explained in the Automatic Image Classifier in the Section \ref{sec411}.

{Food-11 Dataset}:
We used subset of Food-11 dataset with 4 categories are Noodles/Pasta, Rice, Soup, and Vegetable/Fruit and used around 40 images of each class for training the model.
While VGG16 and VGG19 were able to achieve an accuracy of around 94\%, MobileNetV2 gave better accuracy as feature extractor with an accuracy of 99\% on validation data set.

{Kaggle Flowers Dataset}:
We used subset of flowers-recognition dataset from Kaggle. The data consists of images corresponding to five classes: daisy, tulip, rose, sunflower, dandelion. For each class, we take about 50 photos for training/validating of our model. The MobileNetV2 model gave a better accuracy as feature extractor here as well.

\begin{table}[ht!]
\centering
\footnotesize
\caption{Image Classification accuracy with different models as feature extractor}
\vspace{0.1in}
\begin{tabular}{|c|c|c|c|}
\hline Data Set & VGG16 & VGG19 & MobileNetV2 \\
\hline Standford Car Dataset   & 88\% & 87\% & 94.1\% \\
\hline Food-11 Dataset & 94\% & 94\% & 99\% \\
\hline Flowers Dataset & 88\% & 88\% & 93\% \\
\hline
\end{tabular}
\label{tablethree}
\end{table}

As observed in Figure \ref{fig15} and Table \ref{tablethree}, MobileNetV2 gave better accuracy when compared to VGG16, VGG19 and other models. Also, MobileNetV2 is better suited for constraint devices due to its smaller footprint as show in Table \ref{model_size_comparision}.

\begin{table}[!ht]
\centering
\footnotesize
\caption{Model size with different models as feature extractor}
\vspace{0.1in}
\begin{tabular}{|c|c|c|}
\hline Feature Extractor & Uncompressed Size & Compressed Size \\
\cline{0-0} & (MB) & (MB)\\
\hline VGG16   & 58.9 & 14.7 \\
\hline VGG19 & 58.9 & 14.7 \\
\hline MobileNetV2 & 8.9 & 2.3 \\
\hline
\end{tabular}
\label{model_size_comparision}
\end{table}

\subsection{Automatic Audio Classification}
In case of Audio Classification, we tried two approaches:
\begin{itemize}
    \item Using Amplitude vs Time plots of Audio and VGG19 as feature extractor
    \item Using log mel spectrogram and YAMNet as feature extractor.
\end{itemize} 

{VGG19 as feature Extractor}: In this approach, we divide the audio into fixed length audio frames and plotted wave forms of these audios files with y axis as Amplitude and x axis as time. These plotted wave forms were provided as inputs to VGG19 for feature extraction and the extracted features were fed to the dense layer for classification.
But the above simple two dimensional representation of the audio files doesn't capture the features of complex audio files as a result the accuracy of this approach was not good when tested on real world audio data.

\begin{figure}
    \begin{center}
    \includegraphics[scale=0.4]{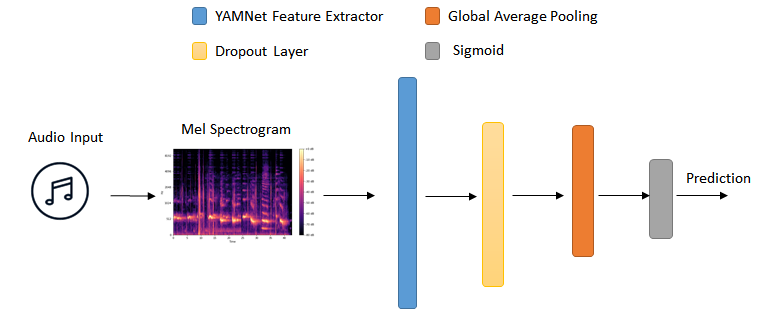}
    \caption{YAMNet as Feature Extractor}
    \label{yamnet_extractor}
    \end{center}
\end{figure}
{YAMNet as Feature Extractor}: YAMNet as a feature extractor gave better accuracy than the above approach using VGG19 model as it uses a stabilized Log Mel Spectrogram as input to the model which is better able to capture the features of audio files as shown in Figure \ref{yamnet_extractor}. Pretrained YAMNet Model is also trained specifically for audio data and with large corpus of AudioSet dataset. Hence, the features extracted by the YAMNet model were able to converge faster and gave better accuracy.

We tested our Automatic Audio Classifier Tool with the following dataset-

{ESC-50 Environment Sounds}: ESC-50 dataset is a labeled collection of 2000 environmental audio recordings. We use a subset of this data set as explained in Section \ref{sec421}. We were able to reach an accuracy of around 98.4\% on validation and test dataset of the model based on YAMNet feature extractor.

{Musical Instrument}: We created a custom dataset of musical instruments consisting of flute, guitar and piano. The training data consists of 50 audio files for each of the 3 classes of length 10 sec. The validation set had 20 audio files for each of the 3 classes. We were able to get an accuracy of 99\% with our automatic model created with YAMNet as feature extractor.

\begin{table}[ht]
\centering
\caption{Audio Classification accuracy with YAMNet}
\vspace{0.1in}
\begin{tabular}{|c|c|c|c|}
\hline Data Set & YAMNet \\
\hline ESC-50 Environment Sounds   & 98.4\% \\
\hline Musical Instruments & 99\% \\
\hline
\end{tabular}
\label{tablefour}
\end{table}

\section{Conclusion and Future Work}
In this paper, we explored a novel approach to create an automatic model creation tool using transfer learning. We described the components involved in our tool and demonstrated an automatic audio and image classifier creation with our tool. 

In case of Image Classification, we used MobileNetV2 as it proved to be better suited for constraint devices and gave better accuracy than VGG16 and VGG19. The extracted features were fed to GlobalAveragePooling2D and a dense layer with softmax activation function for classification. We were able to get an accuracy of $94.1\%$ on a subset of Stanford car dataset using our tool. For Audio Classification, we used YAMNet Model for feature extraction and then fed the extracted feature to GlobalAveragePooling2D and a dense layer with sigmoid activation function for classification. We were able to get an accuracy of $98.4\%$ on the subset of ESC-50 Environment sound dataset. We were also able to achieve a compression of $\sim74\%$ from the standard model size using our tool.

In future, this tool can be extended to create automatic object detection, text classification, activity classification models. Auto data augmentation can be integrated with this tool which can further help to generalize the created model.

\bibliographystyle{IEEEtran}
	\bibliography{transfer_learning}

\end{document}